\documentclass[11pt]{article}
\usepackage[utf8]{inputenc}
\usepackage{bm,amsmath,amssymb,amsthm}
\usepackage{graphicx}
\usepackage{hyperref}
\usepackage{url}
\usepackage{fullpage}
\usepackage{xcolor}
\usepackage{xspace}
\usepackage{booktabs}
\usepackage{multirow}
\usepackage[numbers]{natbib}
\usepackage{amsfonts}
\usepackage{Definitions}
\usepackage{paralist}
\usepackage{subcaption}
\usepackage[normalem]{ulem}
\usepackage{mathtools}
\useunder{\uline}{\ul}{}
\usepackage{enumitem}
\usepackage{algorithm}
\usepackage{algorithmic}
\usepackage{authblk}
\useunder{\uline}{\ul}{}
\definecolor{mydarkblue}{rgb}{0,0.08,0.45}
\hypersetup{
    colorlinks,
    citecolor=mydarkblue,
    filecolor=mydarkblue,
    linkcolor=mydarkblue,
    urlcolor=mydarkblue
}
\graphicspath{{./figs/}}

\begin{document}

\title{\textbf{Variational Reasoning for \\ Question Answering with Knowledge Graph}}
\date{}
\author[1$^*$]{Yuyu Zhang}
\author[1\thanks{Both authors contributed equally to the paper.}]{Hanjun Dai}
\author[2]{Zornitsa Kozareva}
\author[2]{Alexander J. Smola}
\author[1]{Le Song}
\affil[1]{College of Computing, Georgia Institute of Technology}
\affil[2]{Amazon Web Services}
\affil[1]{\texttt{\{yuyu.zhang, hanjun.dai, lsong\}@cc.gatech.edu}} \affil[2]{\texttt{\{kozareva, smola\}@amazon.com}}
\maketitle

\begin{abstract}
Knowledge graph (KG) is known to be helpful for the task of question answering (QA), since it provides well-structured relational information between entities, and allows one to further infer indirect facts. However, it is challenging to build QA systems which can learn to reason over knowledge graphs based on question-answer pairs alone. First, when people ask questions, their expressions are noisy (for example, typos in texts, or variations in pronunciations), which is non-trivial for the QA system to match those mentioned entities to the knowledge graph. Second, many questions require multi-hop logic reasoning over the knowledge graph to retrieve the answers. To address these challenges, we propose a novel and unified deep learning architecture, and an end-to-end variational learning algorithm which can handle noise in questions, and learn multi-hop reasoning simultaneously. Our method achieves state-of-the-art performance on a recent benchmark dataset in the literature. We also derive a series of new benchmark datasets, including questions for multi-hop reasoning, questions paraphrased by neural translation model, and questions in human voice. Our method yields very promising results on all these challenging datasets.
\end{abstract}

\section{Introduction}
Question answering (QA) has been a long-standing research problem in Machine Learning and Artificial Intelligence. Thanks to the creation of large-scale knowledge graphs such as DBPedia~\citep{AueBizKobLeh07} and Freebase~\citep{BolEvaPar08}, QA systems can be armed with well-structured knowledge on specific and open domains. Many traditional approaches for KG-powered QA are based on semantic parsers~\citep{ClaGolCha10, LiaJorKle11, BerChoFro13, YihChaHe15}, which first map a question to formal meaning representation (e.g. logical form) and then translate it to a KG query. The answer to the question can be retrieved by executing the query. One of the disadvantages of these approaches is that the model is not trained end-to-end and errors may be cascaded.

With the recent success of deep learning, some end-to-end solutions based on neural networks have been proposed and show very promising performance on benchmark datasets, such as Memory Networks~\citep{WesChoBor14}, Key-Value Memory Networks~\citep{MilFisDodKaretal16} and Gated Graph Sequence Neural Networks~\citep{LiTarBroZem15}. However, these neural approaches treat the KG as a flattened big table of itemized knowledge records, making it hard to exploit the structure information in the graph and thus weak on logic reasoning. When the answer is not a direct neighbor of the topic entity in question (i.e. there are multiple hops between question and answer entities in the KG), which requires logic reasoning over the KG, the neural approaches usually perform poorly. For instance, it is easy to handle single-hop questions like ``\textit{Who wrote the paper titled ...?}'' by querying itemized knowledge records in triples (\textit{paper\_title}, \textit{authored\_by}, \textit{author\_name}). However, logic reasoning on the KG is required for multi-hop questions such as ``\textit{Who have co-authored papers with ...?}''. With the KG, we start from the mentioned author, and follow $author \xrightarrow{authored} paper \xrightarrow{authored\_by} author$ to find answers. A common remedy is the so-called knowledge graph completion: create new relations for non-neighbor entity pairs in the KG~\citep{SocCheManNg13, DonGabHeiHorLaoMurStrSunZha14, GuuMilLia15}. However, multi-hop reasoning is combinatorial in nature, i.e. the number of multi-hop relations grow explosively with the increase of hops. For example, if we create new relation types like \textit{friend-of-friend} and \mbox{\textit{friend-of-friend-of-friend}}, the number of edges in the KG will explode, which is intractable for both storage and computation.

Another key challenge is how to locate topic entities in the KG. Most existing works assume that the topic entity in question can be located by simple string matching~\citep{MilFisDodKaretal16, DodGanZha15, LiTarBroZem15, BerChoFro13}, which is often not true. When people ask questions, either in text or speech, various noise can be introduced in the expressions. For example, people are likely to make typos or name ambiguity in question. In even harder case, audio questions, people may pronounce the same entity differently in different questions, even for the same person. Due to these noises, it is hard to do exact matching to locate topic entities. For text questions, broad matching techniques (e.g. hand-craft rules, regular expressions, edit distance, etc.) are widely used for entity recognition~\citep{RaoMcnDre13}. However, they require domain experts and lots of human effort. For speech questions, it is even harder to match topic entities directly. Most existing QA systems first do speech recognition, converting the audio to text, and then match entities in text. Unfortunately, the error rate is typically high for speech recognition system to recognize entities in voice, such as human names or street addresses. Since it is not end-to-end, the error of the speech recognition system may cascade to affect the downstream QA system.

Typically, the training data for QA system is provided as question-answer pairs, where fine-grained annotation of these pairs are not available, or only available for a few. More specifically, there are very few explicit annotations of the exact entity present in the question, the type of the questions, and the exact logic reasoning steps along the knowledge graph leading to the answer. Thus it is challenging to simultaneously learn to locate the topic KG entity in the question, and figure out the unknown reasoning steps pointing to the answer based on training question-answer pairs alone.

To address the challenges mentioned above, we propose an end-to-end learning framework for question answering with knowledge graph named variational reasoning network (VRN), which have the following new features:
\begin{itemize}
  \item We build a probabilistic modeling framework for end-to-end QA system, which can simultaneously handle uncertain topic entity and multi-hop reasoning.
  \item We propose a novel propagation-like deep learning architecture over the knowledge graph to perform logic inference in the probabilistic model.
  \item We apply the REINFORCE algorithm with variance reduction technique to make the system end-to-end trainable.
  \item We derive a series of new challenging benchmark datasets \textsc{MetaQA}\footnote{Our new benchmark dataset collections \textsc{MetaQA} are publicly available at \url{https://goo.gl/f3AmcY}.} (MoviE Text Audio QA) intended for research on question-answering systems. These datasets contain over 400K questions for both single- and multi-hop reasoning. To test QA systems in more realistic (and more difficult) scenarios, \textsc{MetaQA} also provides neural-translation-model-paraphrased datasets, and text-to-speech-based audio datasets.
\end{itemize}

Extensive experiments show that our method achieves state-of-the-art performance on both single- and multi-hop datasets, demonstrating the capability of multi-hop reasoning. Moreover, we obtain promising results on the challenging audio QA datasets, showing the effectiveness of end-to-end learning framework. With the rise of virtual assistant tools (e.g. Alexa, Cortana, Google Assistant and Siri), QA systems are now even closer to our daily life. This paper is one step towards more realistic QA systems, which can handle noisy question input in both text and speech, and learn from examples to reason over the knowledge graph.

\section{Related Work}\label{sec:related_work}
\textbf{QA with semantic parser:}
Most traditional approaches for KG-powered QA are based on semantic parsers, which map the question to a certain meaning representation or logical form~\citep{ClaGolCha10, LiaJorKle11, KwiChoArt13, BerChoFro13, YihChaHe15, MarUsbAxeHof14, HofWalMarUsb16}, or directly map the question to an executable program~\citep{LiaBerLeFor16}. These approaches require domain-specific grammars, rules, or fine-grained annotations. Also, they are not designed to handle noisy questions, and do not support end-to-end training since they use separate stages for question parsing and logic reasoning.

\noindent \textbf{Neural approaches for QA:} The family of memory networks achieves state-of-the-art performance in various kinds of QA tasks. Some of them are able to do reasoning within local context~\citep{Kumaretal15, SukWesFeretal115} using attention mechanism~\citep{YanHeGaoDenSmo15}. For QA with KG, \citet{MilFisDodKaretal16} achieves state-of-the-art performance, outperforming previous works~\citep{BorChoWes14, WesChoBor14} on benchmark datasets. Recent work~\citep{NeeLeAbaMccAmo16} uses neural programmer model for QA with single knowledge table. However, the multi-hop reasoning capability of these approaches depends on recurrent attentions and there is no explicit traversal over the KG.

\noindent \textbf{Graph embedding:} Recently, researchers have built deep architectures to embed structured data, such as trees~\citep{Socheretal13, IrsCar14, MouLiZhaWanetal16} or graphs~\citep{DuvMacIpaBometal15, DaiDaiSon16, AtwTow16}. Also some works~\citep{LiTarBroZem15, Johnson17} extend it to sequential case like multi-step reasoning. However, these approaches only work on small instances like sentences or molecules. Instead, our work embeds the reasoning-graph from source entity to every target entity in large-scale knowledge graph. 

\noindent \textbf{Multi-hop reasoning:} There are some other works on knowledge graph completion with traversal, which requires path sampling~\citep{GuuMilLia15, NeeRotMcC15} or dynamic programming~\citep{TouLinYihPooetal16}. Our work can handle QA with natural language or human speech, and the reasoning-graph embeddings can represent complicated reasoning rules. 

In summary, most of the existing approaches have separate stages for entity locating, such as keyword matching, frequency-based method, and domain-specific methods~\citep{YanCha16}. Since they are not jointly trained with the reasoning part, the errors in entity locating (e.g. incorrectly recognized name entity from speech recognition system) will be cascaded to the downstream QA system.

\section{Model}

\begin{figure*}[t]
  \centering
  \includegraphics[width=\textwidth]{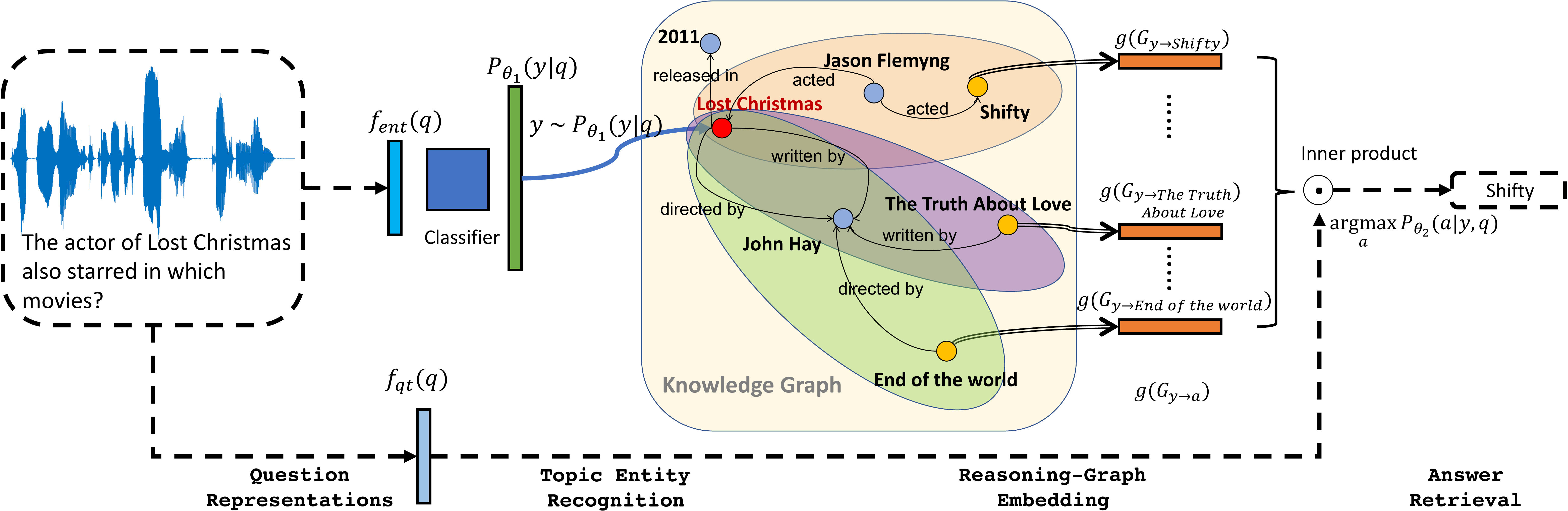}
  \caption{\small End-to-end architecture of the variational reasoning network (VRN) for question-answering with knowledge graph. The model consists of two probabilistic modules for topic entity recognition ($P(y|q)$) and logic reasoning over knowledge graph ($P(a|y,q)$) respectively. Inside the knowledge base plate, the scope of entity \textit{Lost Christmas} (colored red) is illustrated, and each colored ellipsoid plate corresponds to the reasoning graph leading to a potential answer colored in yellow. The reasoning graphs are efficiently embedded and scored against the question embeddings to retrieve the best answer. During training, to handle the non-differentiable sampling operation $y \sim P(y|q)$, we use variational posterior with the REINFORCE algorithm.}
  \label{fig:arch}
\end{figure*}

\subsection{Problem definition}
\label{sec:prob_def}

\noindent \textbf{Knowledge base/graph (KG):} A knowledge graph is a directed graph where the entities and their relations are represented by nodes and edges, respectively, i.e. $\mathcal{G}=(V(\Gcal), E(\Gcal))$. Furthermore, each edge from $E(\Gcal)$ is a triplet $(a_i^{1}, r_i, a_i^{2})$, representing a directed relation $r_i $ between subject entity $a_i^{1}$ and object entity $a_i^{2}$ both from the node set $V(\mathcal{G})$. Each entity in the knowledge graph can also contain additional information such as type and text description. For instance, entity $a_i^{1}$ is described as actor \textit{Jennifer Lawrence}, and entity $a_i^{2}$ is movie \textit{Passengers}. Then a relation in the knowledge graph can be (\textit{Jennifer Lawrence}, \textit{acted\_in}, \textit{Passengers}), where the corresponding $r_i$ is \textit{acted\_in}. In this work, we assume that the knowledge graph is given.

\noindent \textbf{Question answering with KG:}
Given a question $q$, the algorithm is asked to output an entity in the knowledge graph which properly answers the question. For example, $q$ can be a question like ``\textit{who acted in the movie Passengers}?'', and one possible answer is \textit{Jennifer Lawrence}, which is an entity in the KG. In a more challenging setting, $q$ can even be an audio segment reading the same question. The training set $D_{train} = \{(q_i, a_i)\}_{i=1}^N$  contains $N$ pairs of question and answers. 
Note that fine-grained annotation is not present, such as the exact entity present in the question, question type, or the exact logic reasoning steps along the knowledge graph leading to the answer. Thus, a QA system with KG should be able to handle noisy entity in questions and learn multi-hop reasoning directly from question-answer pairs. 

\subsection{Overall formulation}

To address both key challenges in a unified probabilistic framework, we propose the variational reasoning network (VRN). The overall architecture is shown in Fig~\ref{fig:arch}. VRN consists of two probabilistic modules, as described below.

\noindent \textbf{Module for topic entity recognition:} Recognizing the topic entity $y$ (or the entity mentioned in the question) is the first step in performing logic reasoning over the knowledge graph\footnote{In this paper, we consider the case with single topic entity in each question.}. For example, the topic entity mentioned in Sec~\ref{sec:prob_def} is the movie \textit{Passenger}. We denote the topic entity as $y$, and model the compatibility of this entity with the question $q_i$ as a probabilistic model $P_{\theta_1}(y|q_i)$, which shows the probability of the KG entity $y$ being mentioned in the question $q_i$. Depending on the question form (text or audio), the parameterization of $P_{\theta_1}(y|q_i)$ may be different and details can be found in Sec~\ref{sec:ner}.

\noindent \textbf{Module for logic reasoning over knowledge graph:} Given the topic entity $y$ in question $q_i$, one need to reason over the knowledge graph to find out the answer $a_i$. As described in Sec~\ref{sec:prob_def}, the algorithm should learn to use the reasoning rule $(y, \textit{acted\_by}, a_i)$ for that question. Since there is no annotations for such reasoning step, the QA system has to learn it only from question-answer pairs. Thus we model the likelihood of an answer $a_i$ being correct given entity $y$ and question $q_i$ as $P_{\theta_2}(a_i | y, q_i)$. The parameterization of  $P_{\theta_2}(a_i | y, q_i)$ need to capture traversal or reasoning over knowledge graph, which is explained in detail in Sec~\ref{sec:kb_reasoning}.

Since the topic entity in question is not annotated, it is natural to formulate the problem by treating the topic entity $y$ as a latent variable. With the two probabilistic components above, we model the probability of answer $a_i$ being correct given question $q_i$ as $\sum_{y \in V(\Gcal)} P_{\theta_1}(y | q_i) P_{\theta_2}(a_i |  y, q_i)$, which sums out all possibilities of the latent variable. Given a training set $D_{train}$ of $N$ question-answer pairs, the set of parameters $\theta_1$ and $\theta_2$ can be estimated by maximizing the log-likelihood of this latent variable model:
\begin{equation}
\label{eq:mle}
  \max_{\theta_1, \theta_2}~~\frac{1}{N} \sum_{i=1}^N \log\rbr{\sum_{y \in V(\Gcal)} P_{\theta_1}(y | q_i) P_{\theta_2}( a_i |  y, q_i)}.
\end{equation}
Next we will describe our parametrization of $P_{\theta_1}(y | q_i)$ and $P_{\theta_2}(a_i |  y, q_i)$, and the algorithms for learning and inference based on that.

\subsection{Probabilistic module for topic entity recognition \label{sec:ner}}

Most existing QA approaches assume that topic entities are annotated, or can be simply found via string matching. However, for more realistic questions or even audio questions, a more general approach is to build a recognizer that can be trained jointly with the logic reasoning engine.

To handle unlabeled topic entities, we notice that the full context of the question can be helpful. For example, \textit{Michael} could either be the name of a movie or an actor. It is hard to tell which one relates to the question by merely looking at this entity name. However, we should be able to resolve the unique entity by checking the surrounding words in the question. Similarly, in the knowledge graph there could be multiple entities with the same name, but the connected edges (relations) of the entity nodes are different, which helps to resolve the unique entity. For example, as a movie name, \textit{Michael} may be connected with a \textit{directed\_by} edge pointing to an entity of director; while as an actor name, \textit{Michael} may be connected with \textit{birthday} and \textit{height} edges.

Specifically, we use a neural network $f_{\mathrm{ent}}(\cdot): q \mapsto \RR^d$ which can represent the question $q$ in a $d$ dimensional vector. Depending on the question form (text or audio), this neural network can be a simple embedding network mapping bag-of-words to a vector, or a recurrent neural network to embed sentences, or a convolution neural network to embed audio questions. Thus the probability of having $y$ in $q$ is 
\begin{align}
  P_{\theta_1}(y | q) &= \mathrm{softmax}\rbr{W_y^\top f_{\mathrm{ent}}(q)} \\
  &= \frac{\exp(W_y^\top f_{\mathrm{ent}}(q))}{\sum_{y' \in V(\Gcal)} \exp(W_{y'}^\top f_{\mathrm{ent}}(q))}, 
  \label{eq:p_y_given_q}
\end{align}
where $W_y \in \RR^d, \forall y \in V(\Gcal)$ are the weights in the last classification layer. This parameterization avoids heuristic keyword matching for the entity as is done in previous work~\citep{MilFisDodKaretal16, BorChoWes14}, and makes the entity recognition process differentiable and end-to-end trainable.

\subsection{Probabilistic module for logic reasoning over knowledge graph} \label{sec:kb_reasoning}

\begin{algorithm}
	\caption{Joint training of VRN}
	\begin{algorithmic}[1]
		\STATE Initialize $\theta_1, \theta_2, \psi$ with small labeled set
		\FOR{$i=1$ {\bfseries to} $n$}
		\STATE Sample $(q_i, a_i)$ from the training data
		\STATE Sample $\{y_j\}_{j=1}^M$ using \eq{eq:variational_posterior}
		\STATE Smoothing $\tilde{\mu}, \tilde{\sigma}$ with $\{A(y_j, a_i, q_i)\}_{j=1}^M$
		\STATE Update the baseline $b(a, q)$ using least square
		\STATE $\psi \leftarrow \psi - \eta \nabla_{\psi}\Lcal$ using \eq{eq:monte_carlo_grad}
		\STATE $\theta_1 \leftarrow \theta_1 - \eta \nabla_{\theta_1}\Lcal$, $\theta_2 \leftarrow \theta_2 - \eta \nabla_{\theta_2}\Lcal$
		\ENDFOR
	\end{algorithmic}
	\label{algo:train}
\end{algorithm}

\begin{figure}
\centering
\includegraphics[width=0.7\textwidth]{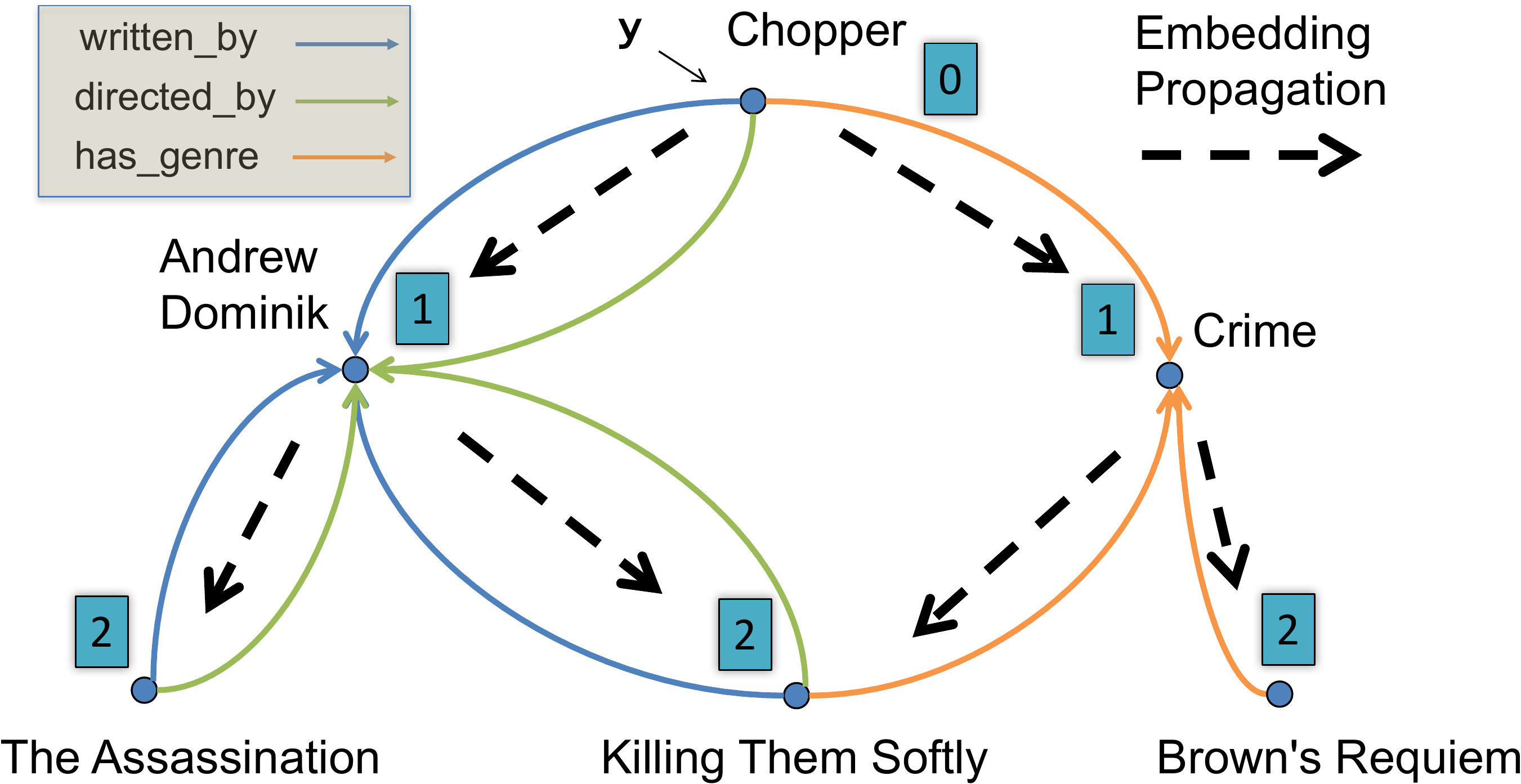}
\caption{A question like ``\textit{movie sharing same genre and director}'' would require two reasoning paths $y \rightarrow$ \textit{Crime} $\rightarrow a$ and $y \rightarrow$ \textit{Andrew Dominik} $\rightarrow a$. The vector representation should encode the information of the entire reasoning-graph, which can be computed recursively. Thus the embedding of \textit{Andrew Dominik} can be reused by \textit{The assassination} and \textit{Killing Them Softly}. }
\label{fig:reasoning_graph}
\end{figure}

Parameterizing the reasoning model $P_{\theta_2}(a | y, q)$ is challenging, since \begin{inparaenum}[1)]
  \item the knowledge graph can be very large;
  \item the required logic reasoning is unknown and can be multi-step. In other words, retrieving the answer requires multi-step traversal over a gigantic graph. 
\end{inparaenum} Thus in this paper, we propose a \textit{reasoning-graph} embedding architecture, where all the inference rules and their complex combinations are represented as nonlinear embeddings in vector space and will be learned. 

\noindent {\bf Scope of $y$.} More specifically, we assume the maximum number of steps (or hops), $T$, of the logic reasoning is known to the algorithm. Starting from a topic entity $y$, we perform topological sort (ignoring the original edge direction) for all entities within $T$ hops according to the knowledge graph. After that, we get an ordered list of entities $a_{1}, a_{2}, \ldots, a_{m}$ and their relations from the knowledge graph. We call this subgraph $\Gcal_{y}$ with ordered nodes as the scope of $y$. Fig~\ref{fig:reasoning_graph} shows an example of a 2-hop scope, where entities are labeled with their topological distance to the source entity. 

\noindent {\bf Reasoning graph to $a$.} Given a potential answer $a$ in the scope $\Gcal_{y}$, we denote $\Gcal_{y\rightarrow a}$ to be the minimum subgraph that contains all the paths from $y$ to $a$ in $\Gcal_y$. The actual logic reasoning leading to answer $a$ for question $q$ is unknown but hidden in the reasoning graph. Thus we will learn a vector representation (or embedding) for $\Gcal_{y\rightarrow a}$, denoted as $g(\Gcal_{y\rightarrow a})\in\RR^d$, for scoring the compatibility of the question type and the hidden path in the reasoning graph.

More specifically, suppose the question is embedded using a neural network $f_{\mathrm{qt}}(\cdot): q \mapsto \RR^d$, which captures the question type and implies the type of logic reasoning we need to perform over knowledge graph. Then the compatibility (or likelihood) of answer $a$ being correct can be computed using the embedded reasoning graph $\Gcal_{y\rightarrow a}$ and the scope $\Gcal_y$ as
\begin{align}
  \label{eq:pa_givenyq}
  P_{\theta_2}(a | y, q) &= \mathrm{softmax}\rbr{f_{\mathrm{qt}}(q) ^\top g(\Gcal_{y\rightarrow a})} \\
  &= \frac{\exp(f_{\mathrm{qt}}(q) ^\top g(\Gcal_{y\rightarrow a}))}{\sum_{a' \in V(\Gcal_y)} \exp(f_{\mathrm{qt}}(q) ^\top g(G_{y\rightarrow a'})) }. 
\end{align}
We note that the normalization in the likelihood requires the embedding of the reasoning graphs for all entities $a^\prime$ in the scope $\Gcal_y$. This may involve thousands of or even more reasoning graphs depending on the KG and the number of hops. Computing these embeddings separately can be very computationally  expensive. Instead, we develop a neural architecture which can compute these embeddings jointly and share intermediate computations.    

\noindent {\bf Joint embedding reasoning graphs.} More specifically, we propose a ``forward graph embedding'' architecture, which is analogous to forward filtering in Hidden Markov Model or Bayesian Network. The embedding of the reasoning graph for $a$ is computed recursively using its parents' embeddings:
\begin{eqnarray}
  \label{eq:forward_embedding}
  g(\Gcal_{y \rightarrow a}) = & \frac{1}{\#\text{Parent}(a)}  \sum_{a_j \in \text{Parent}(a), (a_j, r, a)~\text{or}~(a, r, a_j) \in \Gcal_y} \nonumber \\
  & \sigma( V \times [g(\Gcal_{y \rightarrow a_j}), \vec{e}_r]), 
\end{eqnarray}
where $\vec{e}_r$ is the one-hot encoding of relation type $r \in \Rcal$, $V \in \RR^{d\times (d + |\Rcal|)}$ are the model parameters, $\sigma(\cdot)$ is a nonlinear function such as ReLU, and $\#\text{Parent}(a)$ counts the number of parents of $a$ in $\Gcal_y$. The only boundary case is $g(G_{y \rightarrow y}) = \vec{0}$ when $y = a$. Overall, computing the embedding $g(\Gcal_{y \rightarrow a})$ for all $a$ takes $O(\abr{V(\Gcal_y)} + \abr{E(\Gcal_y)})$ time, which is proportional to the number of nodes and edges in the scope $\Gcal_y$.

This formulation is able to capture various reasoning rules. Take Fig~\ref{fig:reasoning_graph} as an example: the embedding of the entity \textit{Killing Them Softly} sums up the two embeddings propagated from its parents. Thus it tends to match the reasoning paths from the parent entities. Note that this formulation is significantly different from the work in~\citep{DuvMacIpaBometal15,DaiDaiSon16, AtwTow16}, where embedding is computed for each small molecular graph separately. Furthermore, those graph embedding methods often contain iterative processes which visit each nodes multiple times. 

\section{End-to-end Learning}
In this section, we describe the algorithm for learning the parameters in $P_{\theta_1}(y|q)$ and $P_{\theta_2}(a|y,q)$. The overall learning algorithm is described in Algorithm~\ref{algo:train}.

\subsection{Variational method with inverse reasoning-graph embedding}

EM algorithm is often used to learn latent variable models. However, performing exact EM updates for the objective in~\eq{eq:mle} is intractable since the posterior cannot be computed in closed form. Instead, we use variational inference and optimize the negative Helmholtz variational free energy:
\begin{eqnarray}
\label{eq:variational}
	 \max_{\psi, \theta_1, \theta_2} & \mathcal{L}(\psi, \theta_1, \theta_2) = & \frac{1}{N} \sum_{i=1}^N \EE_{Q_{\psi}(y | q_i, a_i)} [ \nonumber \\
	& & \log P_{\theta_1}(y | q_i) + \log P_{\theta_2}(a_i | y, q_i) \nonumber \\
	& & - \log Q_{\psi}(y | q_i, a_i)], 
\end{eqnarray}
where the variational posterior $Q_{\psi}(y | q, a)$ is jointly learned with the model. Note that \eq{eq:variational} is essentially optimizing the lower bound of \eq{eq:mle}. Thus to reduce the approximation error, a powerful set of posterior distributions is necessary. 

\noindent {\bf Variational posterior.} $Q_{\psi}$ computes the likelihood of the topic entity $y$ for a question $q$, with additional information of answer $a$. Thus besides the direct text or acoustic compatibility of $y$ and $q$, we can also introduce logic match with the help of $a$. Similar to the forward propagation architecture used in Sec~\ref{sec:kb_reasoning}, here we can define the scope $\Gcal_a$ for answer $a$, the inverse reasoning graph $G_{a \rightarrow y}$, and the inverse embedding architecture to efficiently compute the embedding $\tilde{g}(G_{a \rightarrow y})$. Finally, the variational posterior consists of two parts:
\begin{equation}
\label{eq:variational_posterior}
  Q_{\psi}(y | q, a) \propto \exp \Big( \tilde{W}_y^\top \tilde{f}_{\mathrm{ent}}(q) + \tilde{f}_{\mathrm{qt}}(q)^\top \tilde{g}(G_{a \rightarrow y}) \Big),
\end{equation}
where the normalization is done over all entities $y^\prime$ in the scope $\Gcal_a$. Furthermore, the embedding operators $\tilde{f}_{\mathrm{ent}}, \tilde{f}_{\mathrm{qt}}$ and parameters $\{\tilde{W}_y\}_{y \in V(\Gcal)}$ are defined in the same way as \eq{eq:pa_givenyq} and \eq{eq:forward_embedding} but with different set of parameters. One can also share the parameter to obtain a  more compact model. 

\subsection{REINFORCE with variance reduction}
\label{sec:reinforce}

Since the latent variable $y$ in the variational objective~\eq{eq:variational} takes discrete values, which is not differentiable with respect to $\psi$, we use the REINFORCE algorithm~\cite{Williams92} with variance reduction~\cite{MniGre14} to tackle this problem.

First, using the likelihood ratio trick, the gradient of $\Lcal$ with respect to posterior parameters $\psi$ can be computed as (for simplicity of notation, we assume that there is only one training instance, i.e., $N=1$):
\begin{eqnarray}
\label{eq:log_ratio}
	\nabla_{\psi} \mathcal{L} = \EE_{Q_{\psi}(y | q, a)} \Big[ \nabla_{\psi} \log Q_{\psi}(y | q, a)\, A(y, q, a) \Big],
\end{eqnarray}
where $A(y, q, a) = \log P_{\theta_1}(y | q) + \log P_{\theta_2}(a | y, q) - \log Q_{\psi}(y | q, a)$ can be treated as the learning signal in policy gradient. 

Second, to reduce the variance of gradient, we center and normalize the signal $A(y, q, a)$ and also subtract a baseline function $b(q, a)$~\cite{MniGre14}. Finally, the gradient in \eq{eq:log_ratio} can be approximated by the Monte Carlo method using $K$ samples of the latent variable from $Q_\psi$:
\begin{eqnarray}
	\nabla_{\psi} \mathcal{L} \approx & \frac{1}{K} \sum_{j=1}^K \nabla_{\psi} \log Q_{\psi}(y_j | q, a) \nonumber \\
	& \rbr{\frac{ (A(y_j, q, a) - \tilde{\mu}) }{\tilde{\sigma}} - b(q, a)},
\label{eq:monte_carlo_grad}
\end{eqnarray}
where $\tilde{\mu}$ and $\tilde{\sigma}$ estimate the mean and standard deviation of $A(y_j, q, a)$ with moving average. $b(q, a)$ is another neural network that fits the expected normalized learning signal. In our experiments, we simply build a two-layer perceptron with concatenated one-hot answer and question features. Here $b(q, a)$ tries to fit $\tilde{A}(y_j, q, a) = \frac{ (A(y_j, q, a) - \tilde{\mu}) }{\tilde{\sigma}}$ by minimizing the square loss. For other parameters $\theta_1$ and $\theta_2$ in $P_{\theta_1}(y|q)$ and $P_{\theta_2}(a|y, q)$ respectively, the gradients are computed in the normal way. 

\section{Inference}

During inference, we are only given the question $q$, and ideally we want to find the answer by computing $\arg\max_{y, a} \log \rbr{P_{\theta_1}(y | q) P_{\theta_2}(a | y, q)}$. However, this computation is quadratic in the number of entities and thus too expensive. Alternatively, we can approximate it via beam search. So we select $k$ candidate entities $y_1, y_2, \ldots, y_k$ with top scores from $P_{\theta_1}(y | q)$, and then the answer is given by 
\begin{equation}
	a^* = \argmax_{a \in \Gcal_y, y \in \{y_1,y_2,\ldots,y_k\}} \log P_{\theta_2}(a | y, q).
\end{equation}
In our experiments, we found that $k=1$ (equivalent as greedy inference) can already achieve good performance.

\section{Experiments}

\begin{table*}
\centering
\caption{Test results (\% hits@1) on Vanilla and Vanilla-EU datasets. EU stands for entity unlabeled.}
\label{table:vanilla}
\resizebox{1.0\textwidth}{!}{%
\begin{tabular}{@{}lcccccc@{}}
\toprule
 & \begin{tabular}[c]{@{}c@{}}Vanilla\\ 1-hop\end{tabular} & \begin{tabular}[c]{@{}c@{}}Vanilla\\ 2-hop\end{tabular} & \begin{tabular}[c]{@{}c@{}}Vanilla\\ 3-hop\end{tabular} & \begin{tabular}[c]{@{}c@{}}Vanilla-EU\\ 1-hop\end{tabular} & \begin{tabular}[c]{@{}c@{}}Vanilla-EU\\ 2-hop\end{tabular} & \begin{tabular}[c]{@{}c@{}}Vanilla-EU\\ 3-hop\end{tabular} \\ \midrule
VRN & \textbf{97.5} & \textbf{89.9} & \textbf{62.5} & \textbf{82.0} & \textbf{75.6} & \textbf{38.3} \\
\citet{BorChoWes14}'s QA system & 95.7 & 81.8 & 28.4 & 39.5 & 38.3 & 26.9 \\
KV-MemNN & 95.8 & 25.1 & 10.1 & 35.8 & 10.3 & 10.5 \\
Supervised embedding & 54.4 & 29.1 & 28.9 & 18.1 & 23.2 & 25.3 \\ \bottomrule
\end{tabular}
}
\end{table*}

\begin{table*}
\centering
\caption{Test results (\% hits@1) on NTM-EU and Audio-EU datasets. EU stands for entity unlabeled.}
\label{table:ntmaudio}
\resizebox{1.0\textwidth}{!}{%
\begin{tabular}{@{}lcccccc@{}}
\toprule
 & \begin{tabular}[c]{@{}c@{}}NTM-EU\\ 1-hop\end{tabular} & \begin{tabular}[c]{@{}c@{}}NTM-EU\\ 2-hop\end{tabular} & \begin{tabular}[c]{@{}c@{}}NTM-EU\\ 3-hop\end{tabular} & \begin{tabular}[c]{@{}c@{}}Audio-EU\\ 1-hop\end{tabular} & \begin{tabular}[c]{@{}c@{}}Audio-EU\\ 2-hop\end{tabular} & \begin{tabular}[c]{@{}c@{}}Audio-EU\\ 3-hop\end{tabular} \\ \midrule
VRN & \textbf{81.3} & \textbf{69.7} & \textbf{38.0} & \textbf{37.0} & \textbf{24.6} & \textbf{21.1} \\
\citet{BorChoWes14}'s QA system & 32.5 & 32.3 & 25.3 & 18.5 & 19.3 & 15.3 \\
KV-MemNN & 33.9 & 8.7 & 10.2 & 4.3 & 7.0 & 15.3 \\
Supervised embedding & 16.1 & 22.8 & 24.2 & 4.1 & 6.1 & 12.1 \\ \bottomrule
\end{tabular}
}
\end{table*}

\subsection{The \textsc{MetaQA} benchmark} \label{subsec:metaqa}

There is an existing public QA dataset named WikiMovies\footnote{It is available at \url{https://research.fb.com/downloads/babi}.}, which consists of question-answer pairs in the domain of movies and provides a medium-sized knowledge graph~\citep{MilFisDodKaretal16}. However, it has several limitations: 1) all questions in it are single-hop, thus it is not able to evaluate the ability of reasoning; 2) there is no noise on the topic entity in question, so it can be easily located in the knowledge graph; 3) it is generated from very limited number of text templates, which is easy to be exploited by models and of limited practical value. Some small datasets like WebQuestions~\citep{BerChoFro13} are mostly for single-hop questions; while WikiTableQuestions~\citep{PasLia15} involves tiny knowledge table for each question, instead of one large-scale knowledge graph shared among all questions. 

Thus in this paper, we introduce a new challenging question-answer benchmark: \textsc{MetaQA} (MoviE Text Audio QA). It contains more than 400K questions for both single and multi-hop reasoning, and provides more realistic text and audio versions. \textsc{MetaQA} serves as a comprehensive extension of WikiMovies. Due to the page limit, we briefly list the datasets included in \textsc{MetaQA} below, and put more details in Appendix~\ref{apdx:dataset}.

\begin{itemize}
	\item \textbf{Vanilla:} We have the original WikiMovies as the Vanilla 1-hop dataset. For multi-hop reasoning, we design 21 types of 2-hop questions and 15 types of 3-hop questions, and generate them by random sampling from a text template pool. Details and question examples are in Appendix~\ref{apdx:samples}.
    \item \textbf{NTM:} Thanks to the recent breakthrough in neural translation models (NTM), we can introduce more variations over the Vanilla datasets. We use a NTM trained by dual learning techniques~\citep{HeXiaQin16} to paraphrase question by first translating it from English to French, and then sample translations back to English with beam search. The questions in the NTM dataset have different wordings but keep the same meaning. This dataset also contains 1-hop, 2-hop and 3-hop categories.
    \item \textbf{Audio:} To make it even more practical and challenging, we generate audio datasets with the help of text-to-speech (TTS) system. We use Google TTS service to read all the questions in Vanilla. We also provide extracted MFCC features for each question. The Audio dataset also contains 1-hop, 2-hop and 3-hop categories. Note that although the audio is machine-generated, it is still much less regulated compared to text-template-generated data, and have a lot of variations in waveforms. For example, even for the same word, the TTS system can have different intonations depending on the word position in question and other context words. Visualization of the audio data can be found in Appendix~\ref{apdx:audio_vis}.
\end{itemize}

\begin{figure}[t]
\centering
\includegraphics[width=0.6\textwidth]{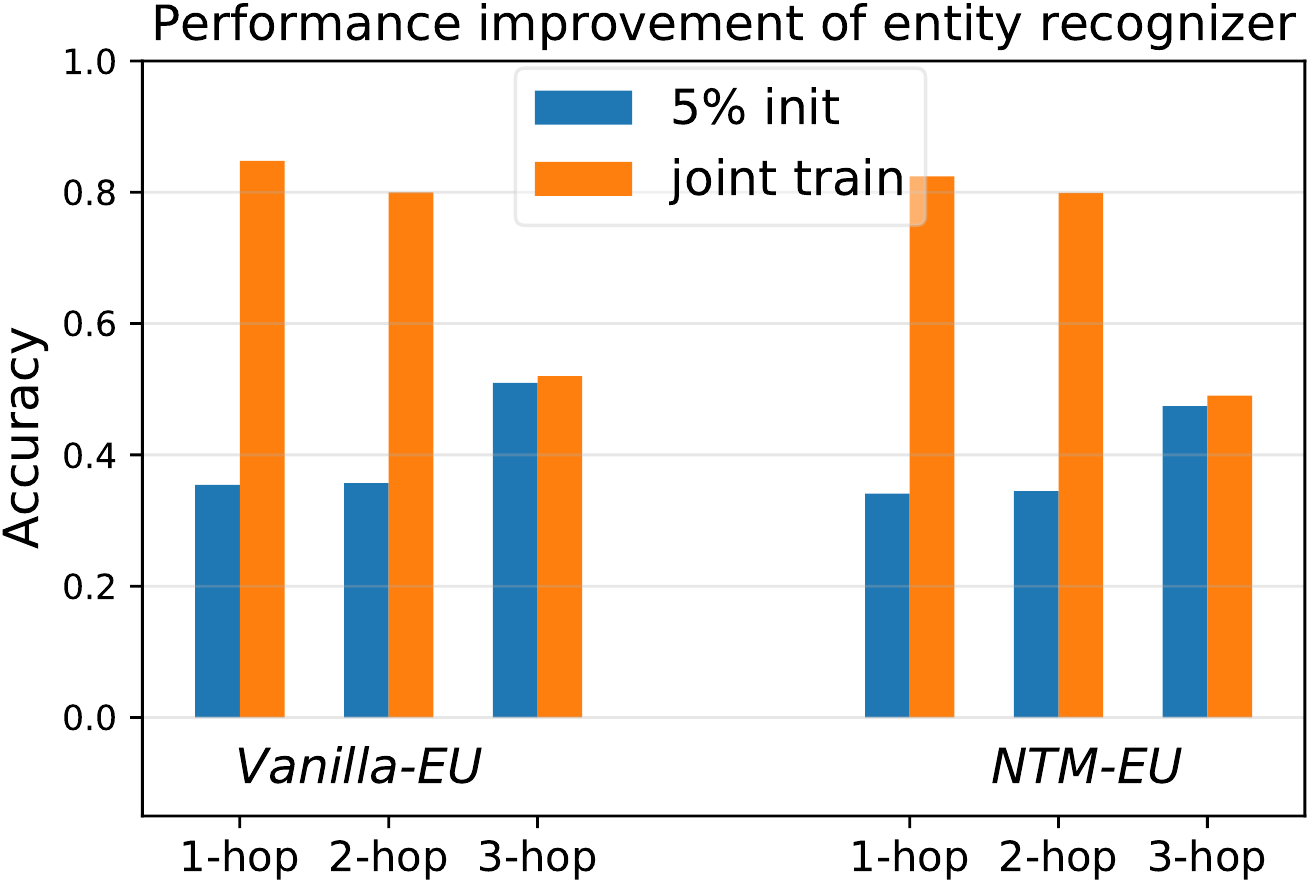}
\caption{Improvement of the entity recognizer. \label{fig:entity_improve} }
\end{figure}

\subsection{Competitor methods}

We have three competitor methods: 
1) as discussed in Sec~\ref{sec:related_work}, \citet{MilFisDodKaretal16} proposed Key-Value Memory Networks (KV-MemNN), and reported state-of-the-art results at that time on WikiMovies;
2) \citet{BorChoWes14}'s QA system also tries to embed the inference subgraph for reasoning, but the representation is simply an unordered bag-of-relationships and neighbor entities;
3) the ``supervised embedding'' is considered as yet another baseline method, which is a simple approach but often works surprisingly well as reported in~\citet{DodGanZha15}.

We implement baseline methods with Tensorflow~\citep{AbaAgaBarBreetal15}.
Our results on Vanilla 1-hop are consistent with the reported performance in ~\cite{MilFisDodKaretal16}. We take whichever higher and report it in Table~\ref{table:vanilla}. For example, our KV-MemNN obtains 95.8\% test accuracy, while the original paper reports 93.9\% on the same dataset, so we just report 95.8\% in table.

When training KV-MemNN, we use the same number of ``internal hops'' as the hop number of that dataset. We also try to use more ``internal hops'' than the dataset hop number, but it is not helpful. Also, we insert knowledge items within 3 hops of the located topic entity to the memory slots, which ensures that if the topic entity is correctly matched, the answer is existing somewhere in the memory array.

\subsection{Experimental settings}

We use all the datasets in \textsc{MetaQA} for experiments. We follow the same split of train/validation/test for all datasets. The number of questions in each part is listed in Appendix (Table~\ref{table:data_stats}). We tune hyperparameters on validation set for all methods. In both Vanilla and NTM, we use bag-of-words representation for entity name to parameterize $W_y$ in \eq{eq:p_y_given_q}.

For Vanilla, we have two different settings: 1) provide the entity labels in all questions, so that we can compare with KV-MemNN under the same setting of \citet{MilFisDodKaretal16} on Vanilla 1-hop dataset; 2) only provide 5\% entity labels among all questions, named as Vanilla-EU (EU stands for topic entity unlabeled). We make all the methods use bag-of-words representation of the question, and avoid hard entity matching. This setting is more of a sanity check of how much the method is dependent on labeled topic entities. In practice, hard matching can always be an option on text data, but it is not feasible for audio data.

To make task more realistic and challenging, we experiment with EU setting for NTM and Audio datasets. For NTM-EU, only 5\% topic entity labels among all questions are provided. For Audio-EU, a higher labeled ratio 20\% since it is much more difficult than text data. To handle the variant length of audio questions, we use a simple convolutional neural network (CNN) with three convolutional layers and three max-pooling layers to embed the audio questions into fixed-dimension vectors. We put more details about CNN embedding in Appendix~\ref{apdx:CNN}.

For all the EU setting above, the small set of entity labeled questions are used to initialize a topic entity recognizer. After that, all methods train on entire dataset but without the entity labels. For VRN, we show that this pretrained recognizer will also get improved with variational joint training; for other baselines, the entity recognizer will be fixed. 

\subsection{Results and discussions}

The experimental results are listed in Table~\ref{table:vanilla} and Table~\ref{table:ntmaudio}.

\noindent \textbf{Vanilla:} Since all the topic entities are labeled, Vanilla mainly evaluates the ability of logic reasoning. Note that Vanilla 1-hop is the same as WikiMovies, which is included for sanity check. All the baseline methods achieve similar performance as reported in the original papers~\citep{MilFisDodKaretal16, BorChoWes14}, while our method performs the best. It is clear to see that 2- and 3-hop questions are harder, leading to significant accuracy drop on all methods. Nevertheless, our method still achieves promising results and lead competitors by a large margin. We notice that KV-MemNN is not performing well on multi-hop reasoning, perhaps due to explosion of relevant knowledge items. 

\noindent \textbf{Vanilla-EU:} Without topic entity labels, all reasoning-based methods are getting worse on multi-hop questions. However, supervised embedding gets better in this case, since it just learns to remember the pair of question and answer entities. According to the statistics in Appendix (Table~\ref{table:new_ratio}), a big portion of questions can be answered by just memorizing the pairs in training data. That explains why supervised embedding behaves differently on this dataset.

\noindent \textbf{NTM-EU:} The questions in this dataset are paraphrased by neural translation model, which increases the variety of wordings, and makes the task harder. It is reasonable that all methods are getting slightly worse results compared to Vanilla-EU. The same explanation applies to supervised embedding, which is not reasoning but memorizing all the pairs. This is indeed weak generalization and it takes advantage of the nature of this dataset, but it is not likely to perform well on new entity pairs.

\noindent \textbf{Audio-EU:} This audio dataset is the most challenging one. As mentioned in Sec~\ref{subsec:metaqa}, even the same word can be pronounced in a variety of intonations. It is hard to recognize the entity in audio data, also hard to tell the question type. It is not surprising that all methods perform worse compared to text data. Our method achieves 37\% on 1-hop audio questions, which is very promising. For 2-hop and 3-hop questions, our method still outperforms other methods. Clearly, there is large room for improvement on audio QA. We leave it as future work, and hopefully the \textsc{MetaQA} benchmark can facilitate more researchers working on QA systems.

\subsection{Model ablation}
\label{sec:entity_recog_improve}

Since our framework uses variational method to jointly learn the entity recognizer and reasoning graph embedding, we here do the model ablation to answer the following two questions: \textit{1) is the reasoning graph embedding approach necessary for inference? 2) is the variational method helpful for joint training?}

\noindent \textbf{Importance of reasoning graph embedding:}
As the results shown in Table~\ref{table:vanilla}, our proposed VRN outperforms all the other baselines, especially in 3-hop setting. Since this experiment only compares the reasoning ability, it clearly shows that simply representing the inference rule as linear combination of reasoning graph entities is not enough. 

\noindent \textbf{Improvement of entity recognition with joint training:}
In Fig~\ref{fig:entity_improve} we show that using our joint training framework with variance reduction REINFORCE, we can improve the entity recognition performance further without the corresponding topic entity label supervision. For 1-hop and 2-hop questions, our model can improve greatly. While for 3-hop, since the inference task is much harder, we can only marginally improve the performance. For audio data, we've improved by 10\% in 1-hop case, and it is hard to improve further for multi hops. In Table~\ref{table:vanilla}, the baselines perform significantly worse in the EU setting, due to the absence of joint training.

\subsection{Inspection of learning and inference}

We study the convergence of our learning algorithm in Appendix~\ref{sec:convergence}. It shows variance reduction technique helps the convergence significantly, while simpler tasks converge better. Also we present an example inference path with highest score in the reasoning graph in Appendix~\ref{sec:vis_inference}. To answer ``\textit{What are the main languages in David Mandel films?}'', the model learns to find the movie \textit{EuroTrip} first through \textit{directed} or \textit{wrote} relationships, then follow \textit{in\_language} to get the correct answer \textit{German}. For visualizing general multi-hop reasoning, attention mechanism in the aggregation operator of each node would be helpful.

\clearpage
\bibliography{main}
\bibliographystyle{unsrtnat}

\newpage
\appendix
\section*{\huge Appendix}

\begin{appendix}
\section{Details of the \textsc{MetaQA} benchmark} \label{apdx:dataset}

\noindent \textbf{Vanilla 1-hop dataset:} Our Vanilla 1-hop dataset is derived from the WikiMovies dataset. Following the settings in \cite{MilFisDodKaretal16}, we use the \texttt{wiki\_entities} branch of WikiMovies. To make it easier to use, we apply an automatic entity labeling on the dataset. Specifically, we parse each question with left-to-right largest consumption of entity names and then normal words, and highlight the entity in question with a pair of square brackets. A few entity names are identical to normal words, which will lead to ``fake entities'' to be labeled in the question. For simplicity, we just remove those ambiguous questions, which makes our Vanilla 1-hop text dataset slightly smaller than WikiMovies. We also provide corresponding question type identifier files for train / validation/ test sets.

\noindent \textbf{Vanilla 2-hop dataset:} The WikiMovies dataset has 1-hop questions only, which inspires us to generate new datasets for 2-hop and 3-hop reasoning. For 2-hop questions, we design 21 question types in total:
\begin{itemize}
\item Actor / Writer / Director to Movie to Actor / Writer / Director / Year / Language / Genre (18 types)
\item Movie to Actor / Writer / Director to Movie (3 types)
\end{itemize}
Following the way that WikiMovies is generated, we design 10 question templates and uniformly randomly sample from them when generating questions.

\noindent \textbf{Vanilla 3-hop dataset:} For 3-hop questions, we don't consider awkward question types: Actor / Writer / Director to Movie to Actor / Writer / Director to Movie, since they're counter-intuitive and quite confusing. For example: The directors of the movies acted by [@entity] have directed which movies? Instead, we construct meaningful 3-hop questions in 15 question types:
\begin{itemize}
\item Movie to Actor / Writer / Director to Movie to Actor / Writer / Director / Year / Language / Genre (different roles for the former and latter roles)
\end{itemize}

\noindent \textbf{NTM datasets:} We use one of the state-of-the-art machine translation models, named dual learning for neural translation model~\cite{HeXiaQin16}, to generate 1-hop, 2-hop and 3-hop NTM datasets. We firstly translate the corresponding Vanilla dataset from English to French, and then translate it back to English with beam search. We guarantee that the topic entity can still be found in the question. By doing so, we can automatically paraphrase the questions, which introduces variations in question wordings and thus leads to more realistic scenario.

\noindent \textbf{Audio datasets:} To generate 1-hop, 2-hop and 3-hop audio datasets, we use Google text-to-speech API~\footnote{We use the API from \url{https://github.com/pndurette/gTTS}.} to read all questions in Vanilla datasets and save the audio as mp3 files. It takes time to process hundreds of thousands of audio files. For convenience, we also provide extracted MFCC features for each question. 

\begin{figure*}
\centering
\begin{subfigure}[t]{0.49\textwidth}
\includegraphics[width=\textwidth]{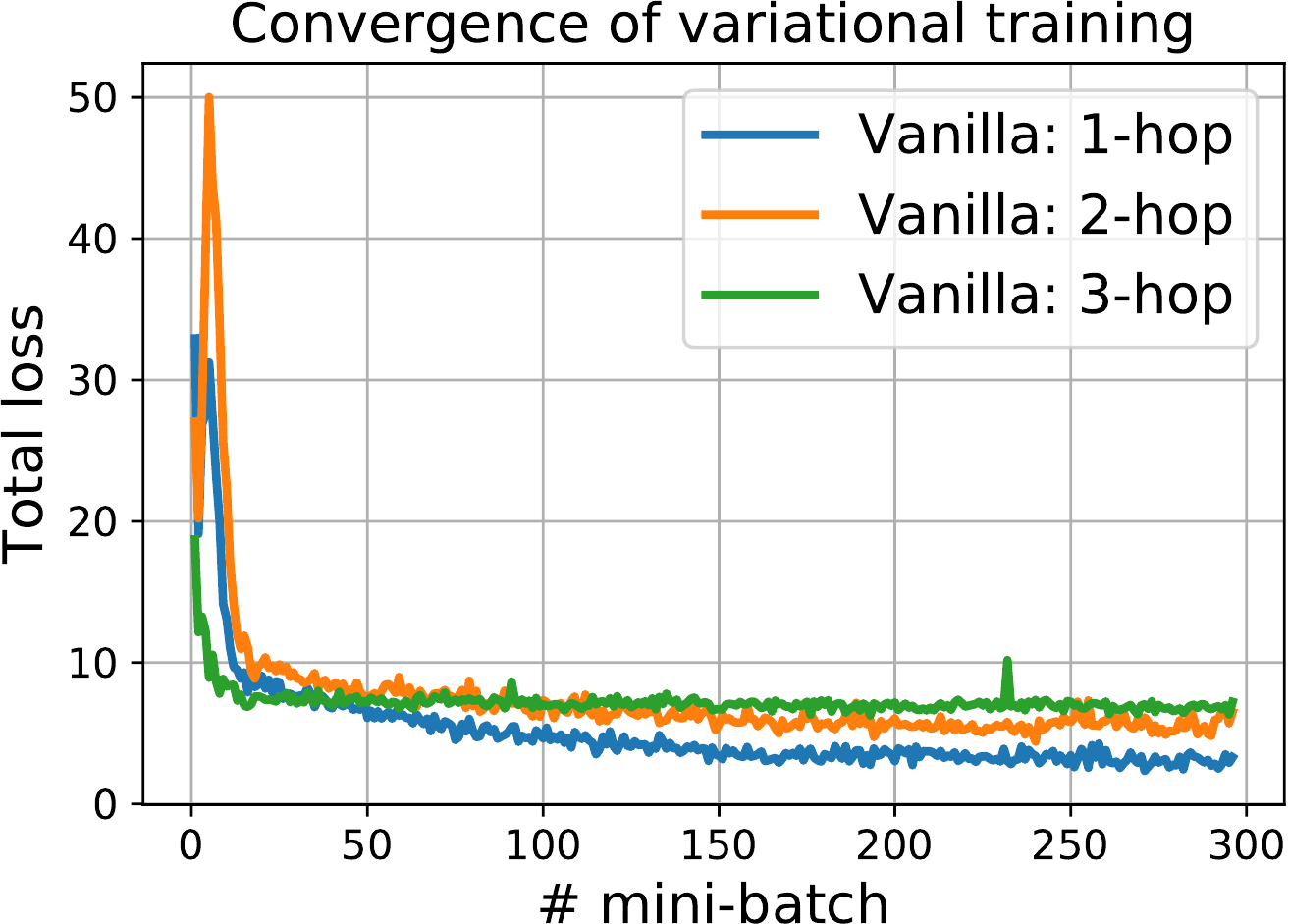}
\caption{Convergence on Vanilla datasets \label{fig:convergence_vanilla}}
\end{subfigure}
\begin{subfigure}[t]{0.49\textwidth}
\includegraphics[width=\textwidth]{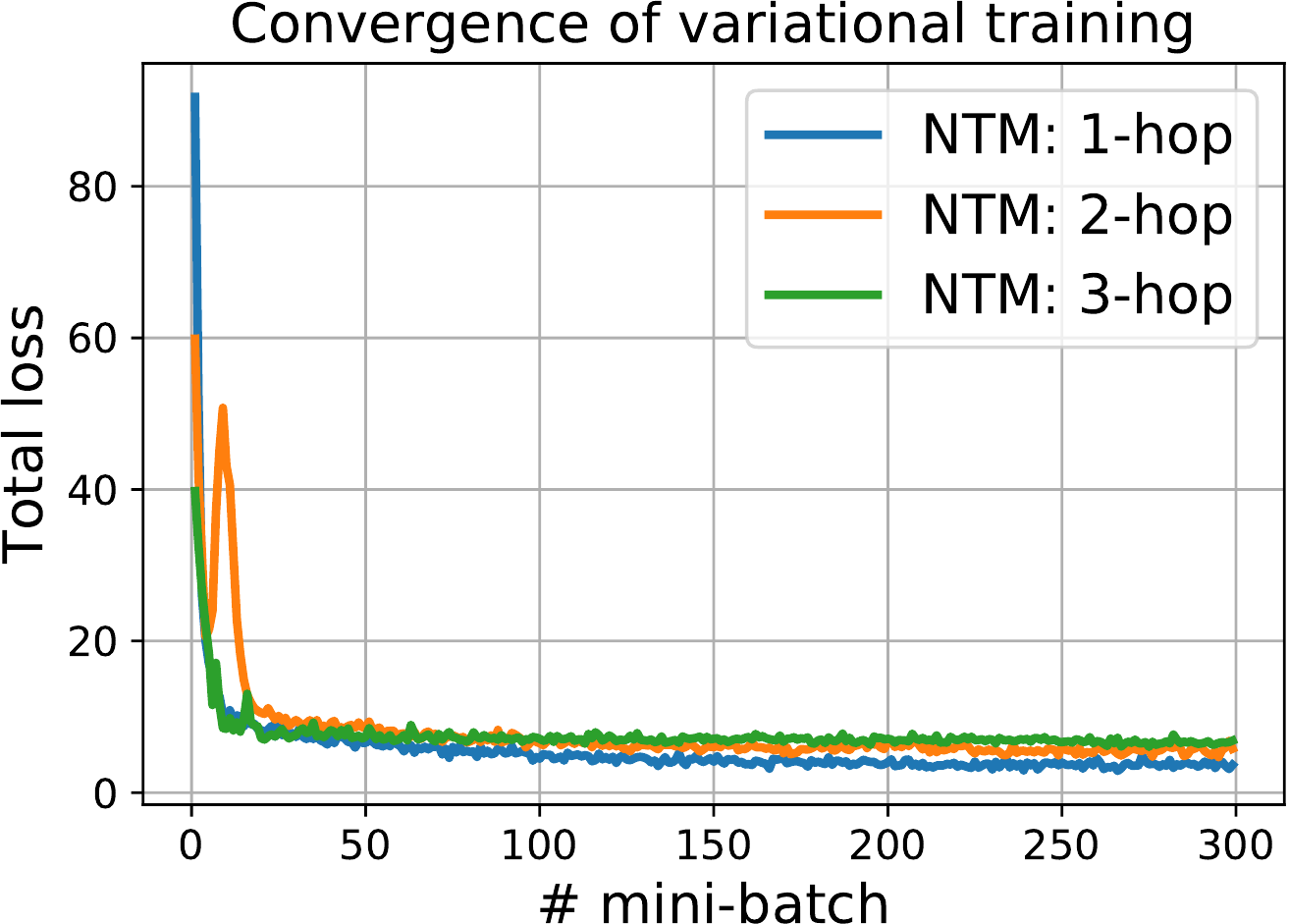}
\caption{Convergence on NTM datasets \label{fig:convergence_ntm} }
\end{subfigure}
\caption{Convergence of our VRN on different text datasets. Here X-axis represents \# mini-batch, and Y-axis reports the total loss for learning $\theta_1, \theta_2, \psi$ and baseline $b(q, a)$.  \label{fig:convergence}}
\end{figure*}

\begin{figure}
\centering
\includegraphics[width=0.7\textwidth]{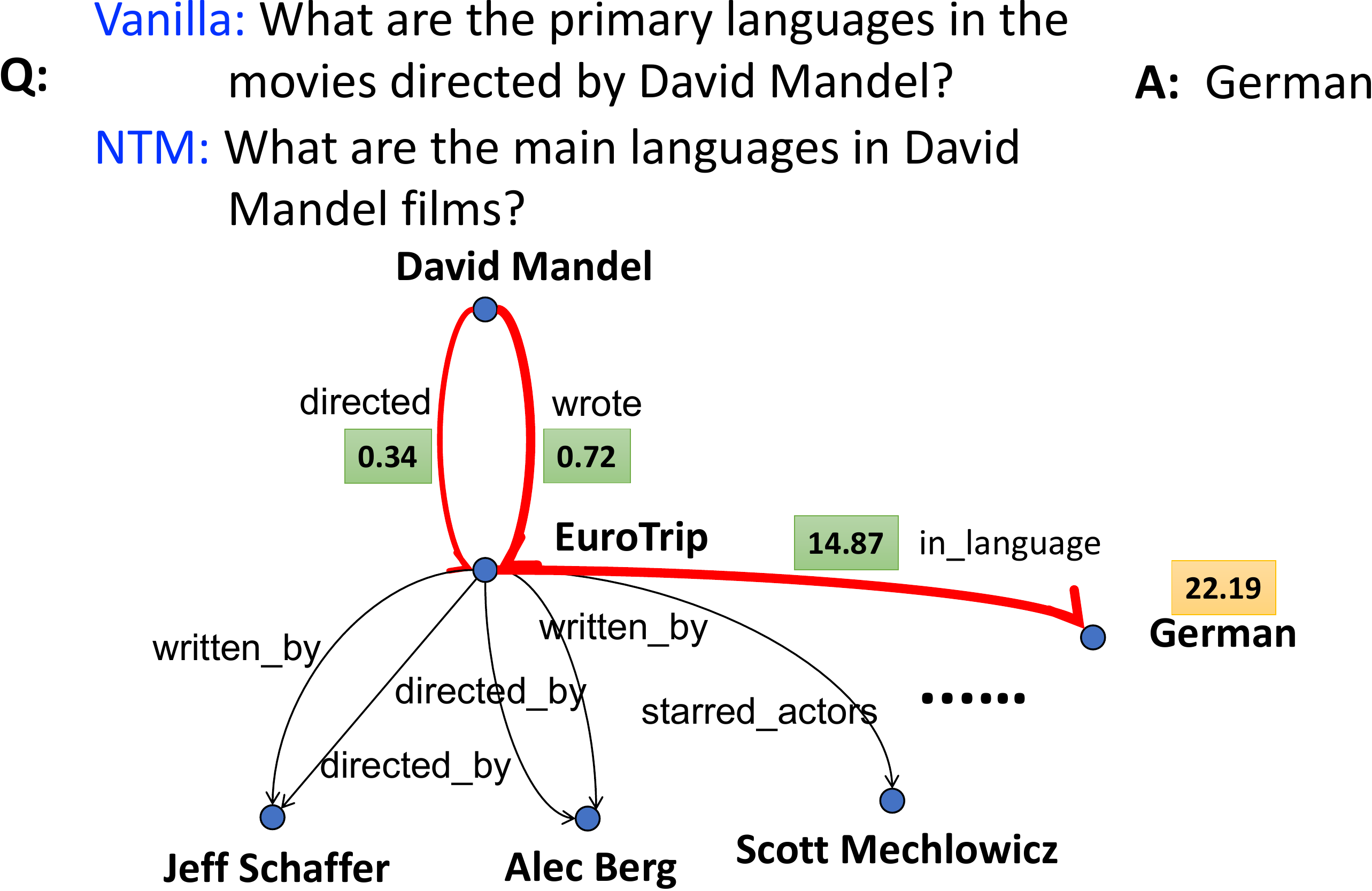}
\caption{Example of the learned 2-hop inference. \label{fig:vis_inference}}
\end{figure}

\begin{table}[t]
\centering
\caption{Statistics of the datasets in the \textsc{MetaQA} benchmark.}
\label{table:data_stats}
\begin{tabular}{@{}lrrr@{}}
\toprule
 & \multicolumn{3}{c}{\# of Questions} \\ \cmidrule(l){2-4} 
 & 1-hop & 2-hop & 3-hop \\ \midrule
Train & 96,106 & 118,980 & 114,196 \\
Validation & 9,992 & 14,872 & 14,274 \\
Test & 9,947 & 14,872 & 14,274 \\ \bottomrule
\end{tabular}
\end{table}

\begin{figure*}[t]
\centering

\begin{subfigure}[t]{0.7\textwidth}
\includegraphics[scale=0.433, trim=139 0 0 0]{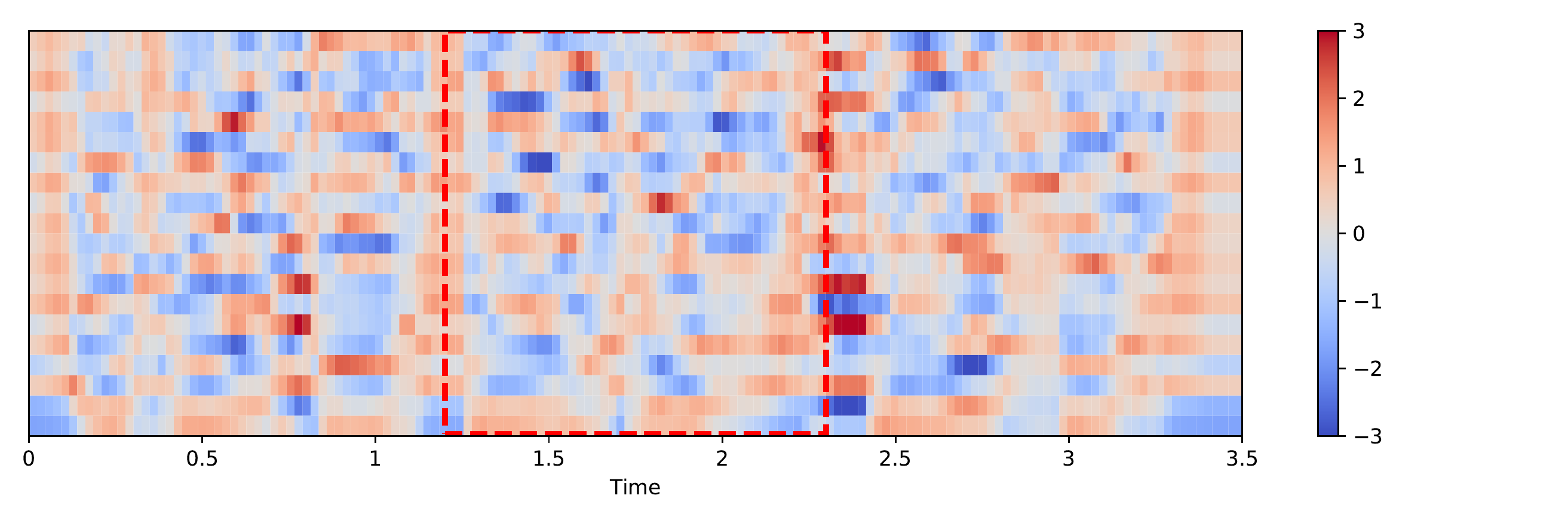}
\caption{MFCC features of question: What films did Dolly Parton star in?
\vspace{15pt}
\label{fig:mfcc1} }
\end{subfigure} 
\begin{subfigure}[t]{0.7\textwidth}
\includegraphics[width=\textwidth]{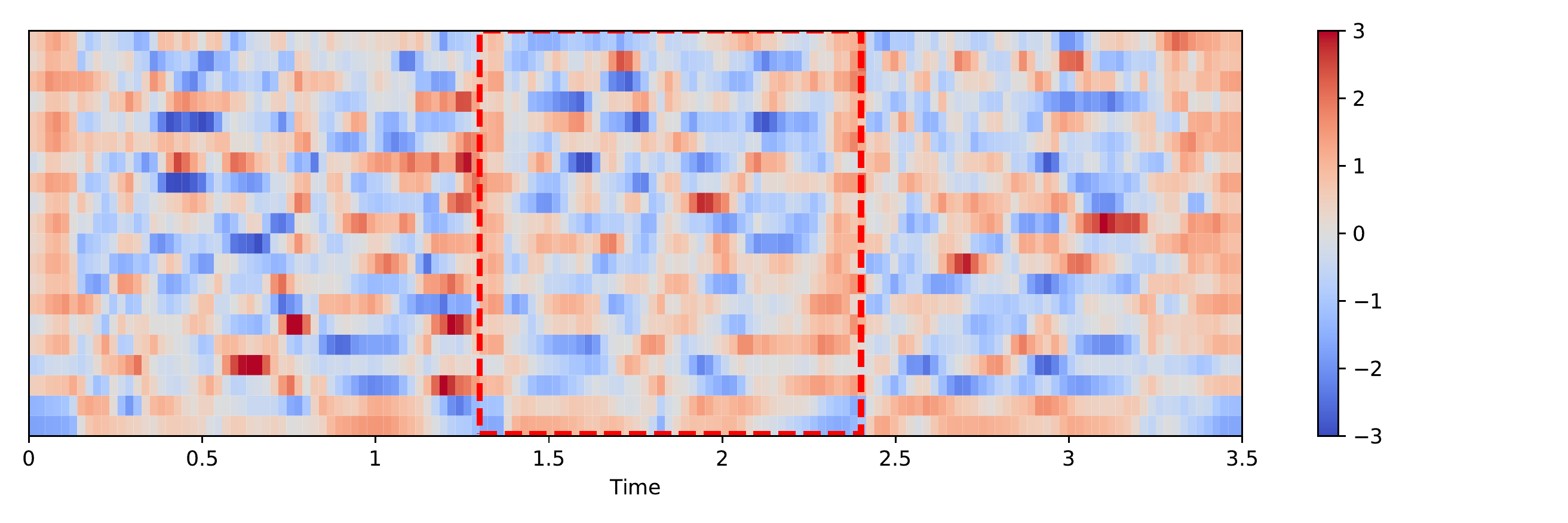}
\caption{MFCC features of question: Which movies was Dolly Parton the writer of? \label{fig:mfcc2} }
\end{subfigure} 
\begin{subfigure}[t]{0.25\textwidth}
\includegraphics[width=\textwidth]{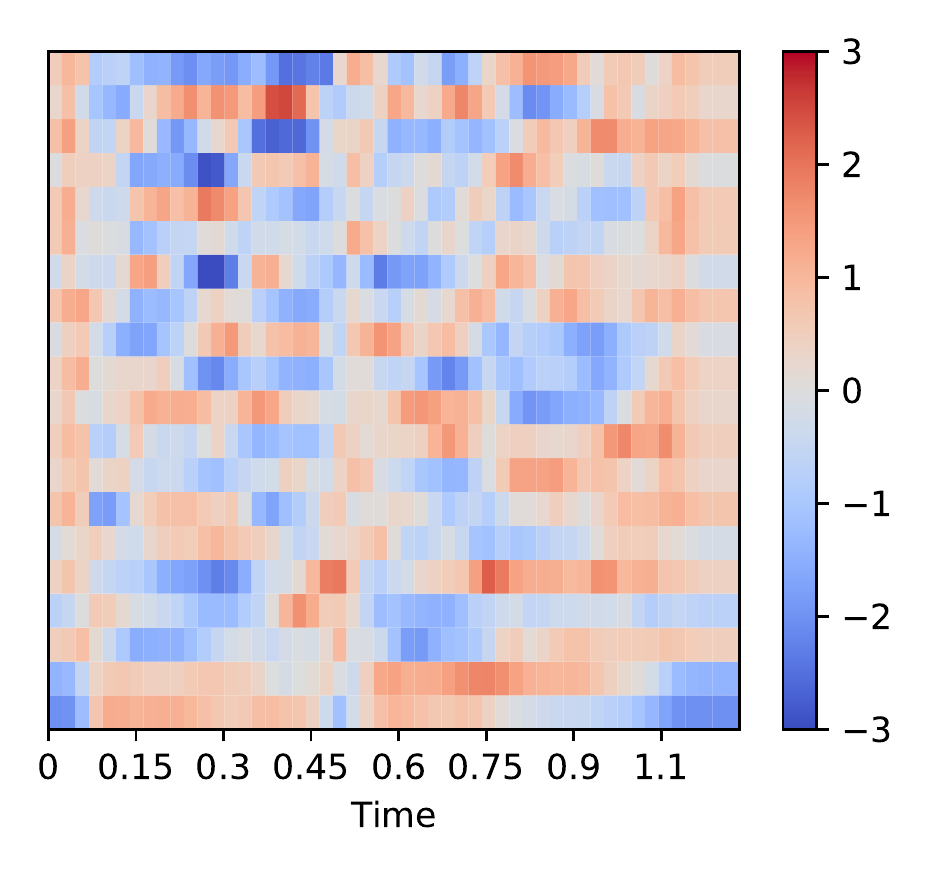}
\caption{MFCC features of the shared entity in both questions: Dolly Parton. \label{fig:mfcc_entity}}
\end{subfigure}
\caption{Visualization of the MFCC features of two questions sharing the same entity. The red dotted box highlights the entity parts. \label{fig:mfcc}}

\end{figure*}

\begin{table}[t]
\centering
\caption{Ratios of new entities and new entity pairs in each dataset.}
\label{table:new_ratio}
\begin{tabular}{@{}lrrr@{}}
\toprule
                                    & 1-hop & 2-hop & 3-hop \\ \midrule
New entities in validation (\%)     & 20.0  & 5.0   & 0.1   \\
New entities in test (\%)           & 20.6  & 4.7   & 0.1   \\
New entity pairs in validation (\%) & 34.1  & 28.1  & 32.1  \\
New entity pairs in test (\%)       & 34.5  & 28.6  & 32.2  \\ \bottomrule
\end{tabular}
\end{table}

\section{Question samples} \label{apdx:samples}
In the \textsc{MetaQA} benchmark, we have in total 21 types of 2-hop questions, and 15 types of 3-hop questions. The exact question types and examples are listed in Table~\ref{table:2hop} and Table~\ref{table:3hop}.

\begin{table*}[t]
\centering
\caption{Examples of the 21 types of 2-hop questions in the \textsc{MetaQA} benchmark.}
\label{table:2hop}
\begin{tabular}{@{}lrp{8cm}@{}}
\toprule
Question Type                 & Count  & Example                                                                \\ \midrule
Movie to Actor to Movie       & 11,709 & The actor of Ruby Cairo also starred in which films?                   \\
Movie to Director to Movie    & 11,412 & Which films share the same director of Vampires Suck?                  \\
Movie to Writer to Movie      & 8,817  & Which movies have the same screenwriter of The Pianist?                \\
Actor to Movie to Actor       & 9,547  & Who co-starred with Joel Evans?                                        \\
Actor to Movie to Director    & 9,241  & Who co-starred with Carlo Ninchi?                                      \\
Actor to Movie to Genre       & 8,548  & What are the genres of the movies acted by Melora Hardin?              \\
Actor to Movie to Language    & 3,067  & What are the main languages in Molly Windsor starred movies            \\
Actor to Movie to Writer      & 8,499  & Who wrote the movies acted by James Madio?                             \\
Actor to Movie to Year        & 10,072 & When did the movies acted by Masato Hagiwara release?                  \\
Director to Movie to Actor    & 4,800  & Who acted in the films directed by Jerry London?                       \\
Director to Movie to Director & 1,797  & Who directed movies together with Chad Stahelski?                      \\
Director to Movie to Genre    & 5,205  & What types are the films directed by Phillip Noyce?                    \\
Director to Movie to Language & 1,850  & What are the languages spoken in the movies directed by Peter Sellers? \\
Director to Movie to Writer   & 3,688  & Who wrote the movies directed by Gary McKendry?                        \\
Director to Movie to Year     & 6,026  & When were the films directed by Jonathan Kahn released?                \\
Writer to Movie to Actor      & 8,447  & Who acted in the films written by Travis Milloy?                       \\
Writer to Movie to Director   & 7,342  & Who directed the films written by Nick Damici?                         \\
Writer to Movie to Genre      & 8,633  & What types are the movies written by Amza Pellea?                      \\
Writer to Movie to Language   & 2,629  & What are the primary languages in the films written by John Musker?    \\
Writer to Movie to Writer     & 7,142  & Who wrote films together with Jonah Hill?                              \\
Writer to Movie to Year       & 10,226 & When did the movies directed by Paul Linke release?                    \\ \bottomrule
\end{tabular}
\end{table*}

\begin{table*}
\centering
\caption{Examples of the 15 types of 3-hop questions in the \textsc{MetaQA} benchmark.}
\label{table:3hop}
\begin{tabular}{@{}lrp{7cm}@{}}
\toprule
Question Type                          & Count  & Example                                                                             \\ \midrule
Movie to Actor to Movie to Director    & 11,600 & Who directed films that share actors with the film Last Passenger?                  \\
Movie to Actor to Movie to Genre       & 11,513 & What types are the films starred by actors in Jack Reacher?                         \\
Movie to Actor to Movie to Language    & 8,735  & What are the languages spoken in the films starred by Blade actors?                 \\
Movie to Actor to Movie to Writer      & 11,516 & Which person wrote the movies starred by the actors in Ludwig?                      \\
Movie to Actor to Movie to Year        & 11,688 & When did the movies starred by Witchboard actors release?                           \\
Movie to Director to Movie to Actor    & 10,784 & Who acted in the films directed by the director of The Road?                        \\
Movie to Director to Movie to Genre    & 10,822 & What types are the films directed by the director of Holly?                         \\
Movie to Director to Movie to Language & 5,909  & The movies that share directors with the movie Effi Briest were in which languages? \\
Movie to Director to Movie to Writer   & 11,005 & Who wrote films that share directors with the film Male and Female?                 \\
Movie to Director to Movie to Year     & 11,350 & When did the films release whose directors also directed Date Movie?                \\
Movie to Writer to Movie to Actor      & 8,216  & Who acted in the movies written by the writer of Bottle Rocket?                     \\
Movie to Writer to Movie to Director   & 8,734  & Who directed films for the writer of Sugar?                                         \\
Movie to Writer to Movie to Genre      & 8,212  & What types are the movies written by the screenwriter of The Gospel?                \\
Movie to Writer to Movie to Language   & 3,908  & What languages are the films that share writers with The Bat in?                    \\
Movie to Writer to Movie to Year       & 8,752  & When did the films release whose screenwriters also wrote Crash?                    \\ \bottomrule
\end{tabular}
\end{table*}

\section{Visualization of audio data} \label{apdx:audio_vis}
We visualize the MFCC features of two questions sharing the same entity, as in Fig~\ref{fig:mfcc}. It is clear that the entity part (highlighted by red dotted lines) is similar but not exactly the same. This shows the difficulty of handling the audio questions.

\section{Details of CNN embedding} \label{apdx:CNN}
To answer audio questions, we need information of both the topic entity in question, and the question type, i.e. what is asking about that entity. So we train two CNNs with different objectives: one is to predict the topic entity, and the other is to predict the question type. We use the same input (MFCC features of audio questions) for both CNNs, and only use training data to fit them. We treat the activations of the second last layer (before softmax layer) as the embeddings of audio questions.

\section{More experiment results}

\subsection{Convergence of VRN}
\label{sec:convergence}

We visualize the training loss in Fig~\ref{fig:convergence} to get an idea of how it converges. We can see using the variance reduction technique, the training converges very fast. Also as expected, for simpler tasks involving fewer inference steps, they can converge to a better solution. 

\subsection{Visualization of the learned reasoning rule}
\label{sec:vis_inference}
To check what the reasoning graph have learned, we visualize the inference path with highest score in the reasoning-graph. Specifically, for 1-hop answers, we simply check the compatibility between edge type and question embedding; for answers with multi-hop, we traverse from answer to topic entity, and take the edge whose embedding has maximum compatibility with the question.

We show one 2-hop inference result in Fig~\ref{fig:vis_inference}. To answer the NTM question correctly, one need to use either \textit{directed} or \textit{wrote} relation to find the movie \textit{EuroTrip}
first, then follow \textit{in\_language} to get the correct answer \textit{German}.

\end{appendix}

\end{document}